\documentclass[letterpaper, 10 pt, conference]{ieeeconf}  

\IEEEoverridecommandlockouts                              

\usepackage{cite}
\usepackage{amsmath,amssymb,amsfonts}
\usepackage[linesnumbered, ruled]{algorithm2e}
\usepackage{graphicx}
\usepackage{textcomp}
\usepackage{xcolor}
\usepackage{verbatim}
\usepackage{subfigure}
\usepackage{multicol}
\usepackage{setspace}
\usepackage{makecell}
\usepackage{url}
\usepackage{array}
\usepackage{caption}
\usepackage[T1]{fontenc}
\usepackage{stfloats}
\usepackage[marginal]{footmisc}
\usepackage{hyperref}
\hypersetup{hypertex=true,
            colorlinks=true,
            linkcolor=blue,
            anchorcolor=blue}
\usepackage{import}
\usepackage{transparent}
\usepackage{color}
\graphicspath{{fig/}}
\title{\LARGE \bf
Robot Navigation in a Crowd by Integrating Deep Reinforcement Learning
and Online Planning}
\author{Zhiqian Zhou, Pengming Zhu, Zhiwen Zeng, Junhao Xiao, Huimin Lu*, Zongtan Zhou
\thanks{*This work was supported by the National Natural Science Foundation of China [61773393, U1913202, U1813205]}
\thanks{All the authors are with Robotics Research Center, College of Intelligence Science and Technology, National University of Defense Technology. zhiqian.zhou13@hotmail.com, zhupengming@nudt.edu.cn, zengzhiwen@nudt.edu.cn, lhmnew@nudt.edu.cn, junhao.xiao@ieee.org, @nudt.edu.cn and narcz@163.com}}
\begin{document}
\maketitle
\thispagestyle{empty}
\pagestyle{empty}
\begin{abstract}
It is still an open and challenging problem for mobile robots navigating along time-efficient and collision-free paths in a crowd. The main challenge comes from the complex and sophisticated interaction mechanism, which requires the robot to understand the crowd and perform proactive and foresighted behaviors. Deep reinforcement learning is a promising solution to this problem. However, most previous learning methods incur a tremendous computational burden. To address these problems, we propose a graph-based deep reinforcement learning method, SG-DQN, that (i) introduces a social attention mechanism to extract an efficient graph representation for the crowd-robot state; (ii) directly evaluates the coarse q-values of the raw state with a learned dueling deep Q network(DQN); and then (iii) refines the coarse q-values via online planning on possible future trajectories. The experimental results indicate that our model can help the robot better understand the crowd and achieve a high success rate of more than 0.99 in the crowd navigation task. Compared against previous state-of-the-art algorithms, our algorithm achieves an equivalent, if not better, performance while requiring less than half of the computational cost.
\end{abstract}
\section{INTRODUCTION}
\indent In the last decade, a significant number of mobile service robots have been introduced into households, offices, and various public places. They share living and social space with humans and have varying degrees of interactions with humans (e.g., carrying foods, cleaning rooms, and guiding visitors). Because humans are dynamic decision-making agents, a robot needs to understand interactions among humans and construct a proactive and foresighted collision avoidance strategy. However, this is still an open and challenging problem.\\
\indent Navigating robots among dynamic obstacles has been thoroughly studied in robotics and related areas. There is a large body of work on classic navigation techniques, such as social force models\cite{sfm_helbing_1995,sfm_ferrer_2013}, and velocity obstacles\cite{vo_fiorini_1998, RVO_van_2008, ORCA_van_2011, brvo_kim_2015}. These approaches often treat humans as simple dynamic obstacles, focusing only on reaction-based collision avoidance rules and ignoring pedestrians' decision-making operations. Therefore, such approaches cannot understand social scenarios well and result in shortsighted and occasionally unnatural behaviors.\\
\begin{figure}[tbp]
\begin{spacing}{0.5}
\end{spacing}
\centerline{\includegraphics[width=1.0\linewidth]{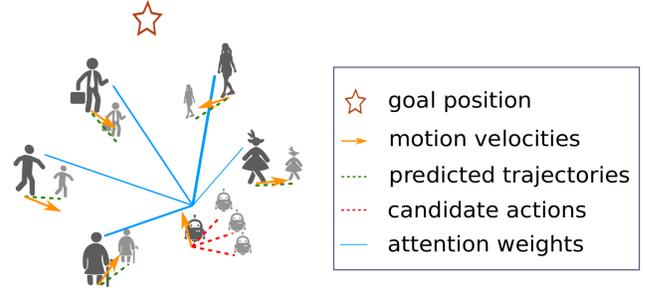}}
\caption{Illustration of our SG-DQN. When the robot navigates in a crowd, it selectively aggregates pedestrians' information with social attention weights, coarsely evaluates the state-action values of all actions with the learned dueling DQN, and quickly generates the best candidate actions. By performing rollouts on the current state, the robot refines the coarse state-action values of the best candidate actions and makes a more foresighted decision.}
\begin{spacing}{0.5}
\end{spacing}
\label{illustration}
\end{figure}  
\indent To address this problem, some researchers separate the crowd navigation task into two disjoint subtasks: first predicting pedestrians' trajectories and then planning collision-free paths \cite{social_2016_alahi, social_2018_vemula, survey_Rudenko_2019, predict_katyal_2020, social_sun_2020, trajectory_tim_2020}. However, with increasing crowd size and density, \textit{trajectory-based} approaches suffer from high computational cost and the \textit{freezing robot problem} \cite{freeze_trautman_2010}, in which the robot cannot find a plausible path. The key to addressing this issue is to consider crowd interactions.\\  
\indent With the rapid development of machine learning algorithms, some other studies have focused on deep reinforcement learning (DRL) because of its ability to model varying interactions in a crowd and generate a more foresighted path. Previous works have focused on value-based and model-based DRL methods\cite{chen_decen_2016,chen_socially_2017, cri_chen_2019, yuying_navigation_2020}. Their framework includes two disjoint modules: a value function to estimate the state value of a given robot-crowd configuration and a state transition function to propagate agents' dynamics forward in time. Furthermore, Chen et al. \cite{rgl_changan_2020} combined the learned model with \textit{Monte-Carlo tree search} (MCTS), selecting the action with maximum d-step return. However, since they must traverse all successor state when estimating the state value, these methods require numerous computational resources, especially when performing a look-ahead rollout. Meanwhile, model-free DRL methods, such as deep Q network (DQN)\cite{mnih-atari-2013, mnih-dqn-2015}, are widely used to learn control policies directly from the raw state with end-to-end training. However, most of them takes raw sensor readings as input (e.g. point clouds and images) \cite{densecavoid_sathy_2020,long_optimally_2018}. This leads to a large state space, which makes training more challenging. Additionally, it is hard to learn a good understanding of a crowd scenario from raw sensor readings, which is the key to a proactive and foresighted navigation policy.\\
\indent A key challenge for learning-based methods in crowd navigation is to get an extensible and efficient state representation, which involves two subproblems. The first subproblem is to model crowd interactions. Most previous models consider only pairwise interactions between the robot and each pedestrian while ignoring human-human interactions in the crowd, which are more common and important \cite{Everett_DRL_2018, everett_collision_2020, yuying_navigation_2020}. LM-SARL, proposed in \cite{cri_chen_2019}, captures human-human interactions via a coarse-grained local map. However, the neighborhoods of humans are $4 \times 4$ grid, which greatly simplifies the human-human interactions in a crowd. The second sub-problem is to aggregate the neighbors' information. The simplest method is to apply a pooling module \cite{chen_decen_2016, socialgan_gupta_2018}, treating all neighbors equally. Another method is to apply a long short-term memory (LSTM) module to combine the pedestrians' information sequentially according to their distances to the robot \cite{Everett_DRL_2018, everett_collision_2020}. Both of them do not consider the crowd as a whole, losing the important structural information. The social attention mechanism\cite{social_2016_alahi, social_2018_vemula}, which models the relative influences among pedestrians in a crowd, is very suitable to selectively aggregate the neighbors' information. In LM-SARL, a social attentive pooling module is built to reason the collective importance of neighboring humans \cite{cri_chen_2019}. Similar work can be found in \cite{yuying_navigation_2020}, where a separate attention network was proposed to infer the relative importance of neighboring humans. After being trained with human gaze data, it is available to accurately predicts human-like attention weights in crowd navigation scenarios.\\
\indent Since a crowd typically produces non-Euclidean data, graph neural networks (GNNs) can be used to extract efficient representations in crowd navigation. In \cite{rgl_changan_2020}, Chen et al. proposed a relational graph model with a two-layer graph convolutional network (GCN) to reason about the relations among all agents \cite{rgl_changan_2020}. However, this algorithm does not show good performance in the crowd navigation task and encounters convergence problems. Even in the given simple simulated environment, the training process is always divergent. Another crowd navigation algorithm, G-GCNRL, utilizes two GCNs to learn human-like attention weights and integrate information about the crowd scenario\cite{yuying_navigation_2020}. The experimental results show that the introduction of GCNs greatly enhances the performance of G-GCNRL.\\
\indent This work focuses on three improvements to address the above issues and proposes a graph-based reinforcement learning method named SG-DQN. First, the social attention mechanism is introduced into the graph attention network (GAT)\cite{gat_velickovic_2018} to extract an efficient graph representation. Second, a dueling DQN \cite{dueling_wang_2016}, is utilized to coarsely evaluate the state-action values, which greatly shortens the time of value estimation. Third, the dueling DQN is combined with online planning, which fine-tunes the coarse evaluation with a simple and rapid rollout performance. Additionally, the reward function is redesigned based on current navigation scenarios to enhance the training convergence. Finally, two types of scenarios, one simple and one complex, are designed to evaluate the proposed approach. The experimental results show that the proposed method helps the robot to better understand the crowd, navigates the robot along time-efficient and collision-free paths, and outperforms state-of-the-art methods. The code and the demonstration video are available at \textit{github.com/nubot-nudt/SG-DQN}.
\section{Problem Formulation}
\subsection{Crowd Navigation Modeling}
As a typical sequential decision-making problem, the crowd navigation task can be formulated in a reinforcement learning framework\cite{chen_decen_2016}. Suppose that a mobile robot navigates in a crowd of N pedestrians over discrete time steps. Natural number $i$ is used to number agents, 0 for the robot and  $i(i>0)$ for the $i$th pedestrian. The agents' configuration is described in robot-centric coordinates, where the robot is located at the origin and the positive direction of the x-axis is from its initial position to its goal position. For each agent, its configuration \textbf{$w$} can be divided into two parts, the observable state and the hidden state. In this work, the observable state consists of the agent's position $\textbf{p}=[p_x, p_y]$, velocity $\textbf{v}=[v_x, v_y]$ and radius \textbf{$\rho$}, and the hidden state includes the agent's intended goal position $\textbf{g}=[g_x, g_y]$, preferred velocity \textbf{$v_{p}$} and heading angle \textbf{$\theta$}. For the robot, it is impossible to observe the pedestrians' hidden states. Therefore, the robot's input state $s$ can be represented as:
\begin{equation}
\begin{aligned}
s&=[w^0,w^1,...,w^N] \\
w^0&=[p_x^0, p_y^0,v_x^0, v_y^0,\rho^0, g_x^0, g_y^0,v_p^0,\theta^0] \\
w^i&=[p_x^i, p_y^i,v_x^i, v_y^i,\rho^i, i>0
\end{aligned}
\end{equation}
where $w^0$ is the full state of the robot and $w^i$ is the observable state of the $i$th pedestrian.\\
\indent At time step $t$, the robot observes a states $s^t$, chooses an action from its discrete action set $a^t \in A$ and then receives an immediate reward signal $r^t=R(s^t,a^t)$ at time step $t+1$. For holonomic robots, the action space consists of the stop action and 80 discrete actions: 5 speeds evenly spaced in (0, $v_p^0$] and 16 headings evenly spaced in [0, 2$\pi$). This work is easy to extend to car-like robots by limiting their headings.
\subsection{Reinforcement Learning Based on the Q-Value}
In this work, a dueling DQN is utilized to directly evaluate the state-action values (q-values for simplicity) and select the best action, which takes the agent-level state as input. There are two main reasons for this. First, the agent-level state can help the robot to better understand crowd scenarios. Second, the dimension of agent-level state is much less than the dimension of raw sensor data, which greatly reduces the computational cost. The objective is to develop the optimal deterministic policy, $\pi^*:s^{t} \rightarrow a^t$, that maximizes the expected discounted return of the robot in reaching its goal:
\begin{equation}
\begin{aligned}
\pi ^*(s^t) =&\mathop{\arg\max}\limits_{a^t}Q^*(s^t,a^t)\\
\end{aligned}
\end{equation}
$Q^*(s,a)$ is the corresponding optimal state-action function, which satisfies the Bellman equation:
\begin{equation}
\begin{aligned}
Q^*(s,a)=\sum_{s',r}P(s',r|s,a)[r+\gamma^{\bigtriangleup t\cdot v_p} \mathop{\max}\limits_{a'}Q^*(s',a')]
\end{aligned}
\end{equation}
where $s'$ and $r$ are the successor state and the immediate reward respectively.  
$\gamma \in (0,1)$ is the discount factor that balances the immediate and future rewards, which is normalized by the preferred velocity $v_p$ and time step $\bigtriangleup t$ \cite{chen_decen_2016, chen_socially_2017}. The transition probability is described by $P(s',r|s,a)$.
\subsection{Reward Shaping}
The reward function is also a highly essential point in deep reinforcement learning. However, previous works ignore this point and apply the reward function from \cite{chen_decen_2016}, which was originally designed to resolve the noncommunicating two-agent collision avoidance problem. As the scene continues to expand, the mismatched reward makes the training process challenging and results in poor training convergence\cite{rgl_changan_2020}. In this work, the reward function is redesigned with three parts: $R_g$, $R_c$ and $R_s$. $R_g$ is designed to navigate the robot moving towards its goal. $R_c$ is built to penalize collision cases, and $R_s$ is designed to reward the robot in maintaining a safe distance from all pedestrians. Formally, the reward function at time step $t$ can be given as:
\begin{equation}
\label{R}
R^t=R_g^t+R_c^t+R_s^t
\end{equation}
where $R_g^t$, $R_c^t$ and $R_s^t$ are given by:
\begin{equation}
\label{rg}
R_g^t=\left\{
\begin{aligned}
&r_{g}  \qquad \qquad  \textnormal{if} \quad ||\textbf{p}_0^t - \textbf{g}_0||<0.2 \\
&0.1(||\textbf{p}_0^{t-1} - \textbf{g}_0||-||\textbf{p}_0^{t} - \textbf{g}_0||) \quad \textnormal{otherwise}
\end{aligned}
\right.
\end{equation}
\begin{equation}
\label{rc}
R_c^t=\left\{
\begin{aligned}
& r_{c} && \textnormal{if} \quad ||\textbf{p}_0^t - \textbf{p}_i^t|| < \rho_0+\rho_i \\
& 0 && \textnormal{otherwise}
\end{aligned}
\right.
\end{equation}
\begin{equation}
\label{rs}
\begin{aligned}
R_s^t&=\sum_{i=1}^{N}f(d^t_i,d_{s}) \\  
f(d^t_i,d_{s})&=\left\{
\begin{aligned}
& \bigtriangleup t \cdot (d^t_i - d_{s})/2 && \textnormal{if} \quad d^t_i<0.2 \\
& 0 && \textnormal{otherwise}
\end{aligned} 
\right. 
\end{aligned}
\end{equation}
Here, $\textbf{p}_0$ and $\textbf{g}_0 $ are the position and goal of the robot, $d_{s}$ denotes the threshold distance that the robot needs to maintain from pedestrians at all time and $d^t_i$ is the actual minimum separation distance between the robot and the $i$th pedestrian during the time step. Here, $r_{g} =10$, $r_{c}=-2.5$ and $d_{s}=0.2$. 
\section{Methodology}
\label{method}
As shown in Fig. \ref{framework}, a new framework is proposed to navigate social robots in a crowd, which can be divided into three parts. The first part is a two-layer GAT that extracts an efficient graph representation from the crowd-robot state, which is the basis of q-value estimators. The second part is the dueling DQN, which coarsely estimates the q-values of the current state. The third part is the online planner, which performs rollouts in the near and short-term future based on the dueling DQN and the environment model. 
\subsection{Graph Representation with Social Attention}
In this work, a social attention mechanism is introduced into the GAT to learn a good graph representation of the crowd-robot state. Unlike \cite{cri_chen_2019}, attention weights are computed for all agents, regardless of the robot or pedestrians. Both robot-human interactions and human-human interactions can be modeled via the same graph convolution operation.\\
\indent In the graph representation, the nodes and edges represent agents and connections between agents. Since agents often have different types, the nodes' dimensions are always different (e.g., $s_0\in{\mathbb{R}^9}$ and $s_i\in{\mathbb{R}^5},i>0$). Therefore, multilayer perceptrons (MLPs) are utilized to extract the fixed-length latent state, which will be fed into subsequent graph convolutional layers. The latent state of the $i$th node is given as:
\begin{figure}[tbp]
\centerline{\includegraphics[width=0.95\linewidth]{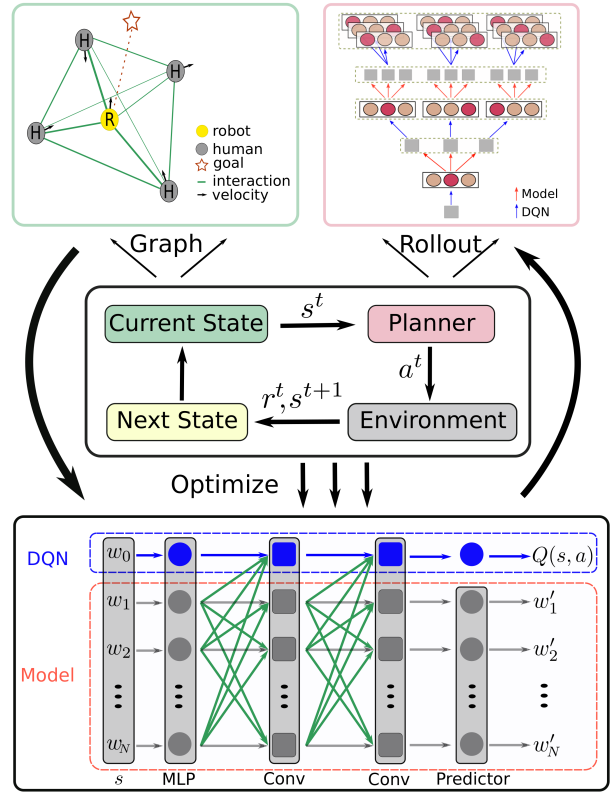}}
\caption{The framework of SG-DQN. The crowd-robot state is described in the form of graph data. Then, the dueling DQN utilizes a two-layer GAT to extract a high-level representation and directly evaluates the q-values of the current state. The core of the online planner is the rollout performance based on the learned dueling DQN and the environment model, both of which are optimized with simulated trajectories.}
\label{framework}
\end{figure} 
\begin{equation}
\label{mlp}
\begin{aligned}
h_0 &=\Psi_r(w_0;W_r)  \\
h_i &=\Psi_p(w_i;W_p), \quad  i=1,2,...,N.  \\
\end{aligned}
\end{equation} 
where $\Psi_r$ and $\Psi_p$ are MLPs with ReLU activations, and their network weights are represented by $W_r$ and $W_p$, respectively. The superscript $l$ is used to denote the layer number. The input of the first layer is denoted by $H^0=[h^0_0,h^0_1,\cdots,h^0_N]$. For all nodes, $h^0_i=h_i$.\\
\begin{figure}[htbp]
\centerline{\includegraphics[scale=0.45]{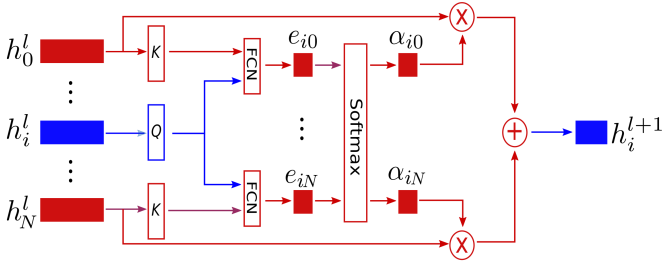}}
\caption{The layerwise graph convolution operation on the $i$th node and its neighborhood (e.g., $h_0$ and $h_N$). $l$ is the layer number, $Q$ is the query matrix and $K$ is the key matrix.}
\label{graph convolution}
\end{figure}
\indent Then, a social attention mechanism is introduced into the spatial graph convolution operation to model interations in a crowd. First, a query matrix $Q$ and a key matrix $K$ are built to transform the input features into higher-level features. Then, these features are concatenated to compute attention coefficients $e_{ij}$ via a fully connected network (FCN). The FCN is followed by a LeakyReLU nonlinearity with a negative input slope $\alpha=0.2$. Finally, the attention coefficients are normalized via a softmax function. A sketch of layerwise graph convolution operation on the $i$th node and its neighborhood is shown in Fig. \ref{graph convolution} and the corresponding layerwise graph convolution rule is given as:
\begin{equation}
\label{attention_equ}
\begin{aligned} 
e_{ij} &= \textnormal{LeakyRuLU}(a(q_i\,||\,k_j)) \\
\alpha_{ij} & = \textnormal{softmax}_j(e_{ij}) =\frac{\exp(e_{ij})}{\sum_{k\in{N(i)}}\exp(e_{ik})}\\
h_i^{l+1} &= \sigma(\sum_{j\in N(i)}\alpha_{ij}^lh_j^l)
\end{aligned}
\end{equation}
where $q_i=\Psi_q(h_i;Q)$, $k_j=\Psi_k(h_j,K)$, $a(\cdot)$ is the attention function and $||$ is the concatenation operation. A layerwise FCN $\Psi_a(\cdot;A)$ is utilized to map concatenated states to attention weights. $\alpha_{ij}$ is the normalized attention weight, indicating the importance of the $j$ node to the $i$th node. $N(i)$ is the neighborhood of the $i$th node in the graph. Similar to previous work\cite{chen_decen_2016, chen_socially_2017, cri_chen_2019, yuying_navigation_2020, rgl_changan_2020}, all agents can obtain accurate observable states for other pedestrians. Therefore, the neighborhood of the $i$th agent includes all pedestrians.\\
\indent Considering that there are indirect interactions in a crowd, the GAT is equipped with two graph convolutional layers to model both direct and indirect interactions in a crowd. The final output of the two-layer GAT can be described by:
\begin{equation}
\label{Hdef}
H=h^0+h^1+h^2
\end{equation} 
\subsection{Graph-Based Deep Q-learning} 
Encouraged by the great success of DQNs \cite{mnih-dqn-2015,mnih-atari-2013} and their variants in reinforcement learning, a dueling DQN is built to estimate q-values in this work. The significant difference from the previous state value network is that the dueling DQN requires only the current state as input. It is a model-free RL method and does not need to evaluate subsequent states.\\
\indent As shown in the bottom of Fig. \ref{framework}, the dueling DQN includes three separate modules: MLPs that extract fixed-length latent states, a two-layer GAT to obtain graph representations $H$, and a dueling architecture to estimate q-values. All these parameters are trained in an end-to-end manner.\\
\indent The dueling architecture consists of two streams that represent the state value and state-dependent advantage functions \cite{dueling_wang_2016}. In this work, a two-layer MLP, $\Psi_c(H;\alpha)$, is built as a common convolutional feature learning module, and two fully connected layers, $\Psi_v(\cdot;\beta)$ and $\Psi_d(\cdot;\eta)$, to obtain a scalar $V(H;\alpha,\beta)$ and an $|A|$-dimensional vector $D(H,a;\alpha,\eta)$. Here, $H$ is the final graph representation mentioned in Eq. \ref{Hdef}, and $\alpha$, $\beta$ and $\eta$ are the parameters of the dueling architecture. Finally, the state-action function $Q(H,a;\cdot)$ can be described by\\
\begin{equation}
\label{q_value}
\begin{aligned}
Q(H,a;\alpha,\beta,\eta)=V(H;\alpha,\beta)+D(H,a;\alpha,\eta)\\
\end{aligned}
\end{equation} 
\begin{algorithm}[tbp]
\caption{Deep Q-learning}\label{q-learning}
\KwOut{Q-value network: $Q(\cdot \,;\theta)$}
Initialize empty experience replay memory $E$\\
Initialize target dueling DQN with random weights $\theta$\\
Duplicate DQN $Q'(\cdot \, ; \theta') \leftarrow Q(\cdot \,; \theta)$ \\
\For{ \textit{$episode \leftarrow 1$} \KwTo \textit{num of episodes}}{
	Initialize random crowd-robot state $S^0$ \\
	\While{\textit{not reach goal, collide or timeout}}
	{$a^t \leftarrow \mathop{\arg\max}\limits_{a^t} Q(s^t,a^t;\theta)$\\
	Execute $a^t$ and obtain $r^t$ and $s^{t+1}$\\
	$s^t \leftarrow s^{t+1}$\\
	Assimilate tuple $(s^t, a^t,r^t, s^{t+1})$ into $E$ \\
	}
	Sample randomly subset $M$ from $E$ \\
	Update $Q(\cdot \,;\theta)$ with $M$ and $Q'(\cdot \, ; \theta')$\\
	\For{every $C$ episodes}{
	evaluate value network $Q(\cdot \,; \theta)$\\
	$Q'(\cdot \, ; \theta') \leftarrow Q(\cdot \,; \theta)$
	}
}
\end{algorithm}
\indent The learning algorithm is shown in Algorithm \ref{q-learning}, where $\theta$ denotes all parameters of the dueling DQN. In each episode, the crowd-robot state $s^0$ is initialized randomly (line 5). Afterward, the robot is simulated to navigate in the crowd using a $\sigma$ greedy policy (line 7). The experience replay technique\cite{Lin_93,mnih-dqn-2015} is also used to increase data efficiency and reduce the correlation among samples. At every step, the newly generated state-value pairs are assimilated into an experience replay memory $E$ (line 10) with a size of $10^5$. In every training, minibatch experiences $M$ are sampled from $E$ to update the network (line 12). To promote convergence, the network $Q(\cdot \,;\theta)$ is cloned to build a target network $Q'(\cdot \, ; \theta')$ (line 16), which generates the Q-learning target values for the next $C=500$ episodes. The loss function is defined as:
\begin{equation}
\label{lossfunction}
\begin{aligned}
L(\theta)&=\sum(r+\gamma^{\bigtriangleup t\cdot v_p} \max \limits_{a'}Q'(s',a';\theta')-Q(s,a;\theta))^2
\end{aligned}
\end{equation}
\subsection{Online Planning Based on Rollout Performance}
\begin{figure}[tbp]
\centerline{\includegraphics[scale=0.35]{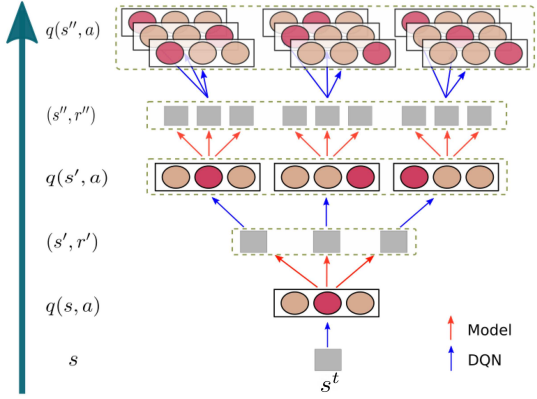}}
\caption{A two-step rollout on state $s^t$. For simplicity, a blue arrow and a red arrow are used to denote the learned DQN and the environment model, respectively.}
\label{rollout}
\end{figure}

\indent Previous works have proven that a DQN has the potential to partly address the shortsightedness problem in crowd navigation. However, with unknown human behavior policies, it is difficult to learn a perfect network model. Inspired by \cite{oh_nips_2015, oh_value_2017, silver2017mastering, schrittwieser_mastering_2020}, the learned DQN is combined with online planning to refine coarse q-values by performing rollouts on the current state and reasoning about the coming future. The depth and width of the rollout performance are denoted by $d$ and $k$. A smaller $k$ means that the navigation policy relies more on model-free RL and has a lower computational burden. When $k=1$, the algorithm degenerates into a pure model-free RL policy. $d$ is designed to balance the importance of the current state and future states, which has a similar effect on the navigation policy as $k$. A two-step rollout in the look-ahead tree search diagram is shown in Fig. \ref{rollout}.\\
\indent In the rollout performance, the learned dueling DQN and the environment model are applied recursively to build a look-ahead tree. In particular, at each expansion, the dueling DQN is used to generate $q$-values and generate the best candidate actions (e.g., actions with the top $k$ $q$-values). The environment model is utilized to predict the corresponding rewards and subsequent states based on the current states and candidate actions. Then, it is possible to refine the $q$-values with $d$-step returns, which can be described as:
\begin{equation}
\label{q_d_value}
Q^d(s,a;\theta)=\left\{
\begin{aligned}
& Q(s,a;\theta) && \textnormal{if} \, d=0 \\
& \frac{d}{d+1}Q(s,a;\theta) + \frac{1}{d+1}(r^{d-1} &&\\
& +\gamma^{\bigtriangleup t\cdot v_p} \max\limits_{a'}Q^{d-1}(s',a';\theta)) && \textnormal{otherwise}
\end{aligned}
\right.
\end{equation}
\indent In addition, the environment model and the RL model are separated here, the environment model can be any model for crowd state prediction or human trajectory prediction. In this work, the crowd prediction framework proposed in \cite{rgl_changan_2020} is used to learn a environment model from simulted trajectories. There are three main reasons. First, the environment model is necessary to perform rollouts. However, it is impossible and impractical for the robot to query the real environment model. Second, with  $R_s$ encouraging the robot to keep a safe distance from pedestrians, model fidelity is less important in the rollout performance. Third, crowd prediction has shown good performance in previous work.
\subsection{Implementation Details}
In this work, the hidden units of $\Psi_r(\cdot)$, $\Psi_h(\cdot)$, $\Psi_r$, $\Psi_v$ and $\Psi_d$ have dimensions (64,32), (64,32), (128), (128) and (128). In every layer, the output dimensions of $\Psi_q(\cdot)$ and $\Psi_k(\cdot)$ are 32. In the dueling architecture, $\Psi_r$,  $\Psi_v$ and $\Psi_d$ have output dimensions of 128, 1, and 81. All the parameters are trained via reinforcement learning and updated by Adam \cite{adam_2015}. The learning rate and discount factor are 0.0005 and 0.9 respectively. In addition, the exploration rate of $\epsilon$-greedy policy decays from 0.5 to 0.1 linearly in the first 5000 episodes and remains 0.1 in the remaining 5000 episodes. In the rollout performance, the planning depth and planning width are set to 1 and 10, respectively.
\section{Experiments}
\subsection{Simulation Setup}
SG-DQN has been evaluated in both simple and complex scenarios. The simple scenario is built based on the CrowdNav\cite{cri_chen_2019}, in which only five pedestrians in a circle of radius 4 $m$ are controlled by ORCA\cite{ORCA_van_2011}. For each pedestrian, his initial position and his goal position are set on the circle, symmetric about the center of the circle. Therefore, they will likely interact near the center of the circle at approximately 4.0 $s$. In addition to the five circle-crossing pedestrians, the complex scenario introduces another five square-crossing pedestrians, whose initial positions and goal positions are sampled randomly in a square with a side length of 10 $m$. The square shares the same center as the circle. In either the simple scenario or the complex scenario, once the pedestrian arrives at his goal position, a new goal position will be reset randomly within the square, and the former goal position is recorded as a turning position. The initial position, the turning position, and the final goal position of pedestrian 0 are marked in Fig. \ref{sim_traj}. Note that the robot is invisible in both scenarios. As a result, the simulated pedestrians will never give way to the robot, which requires the robot to have a more proactive and foresighted collision avoidance policy.
\subsection{Qualitative Evaluation}
\subsubsection{Training Process}
\begin{figure}[tbp] 
	\subfigure[Trainning in simple scenarios]{
	    \begin{minipage}[t]{0.46\linewidth}
	    \centering
		\includegraphics[width=1.0\linewidth ]{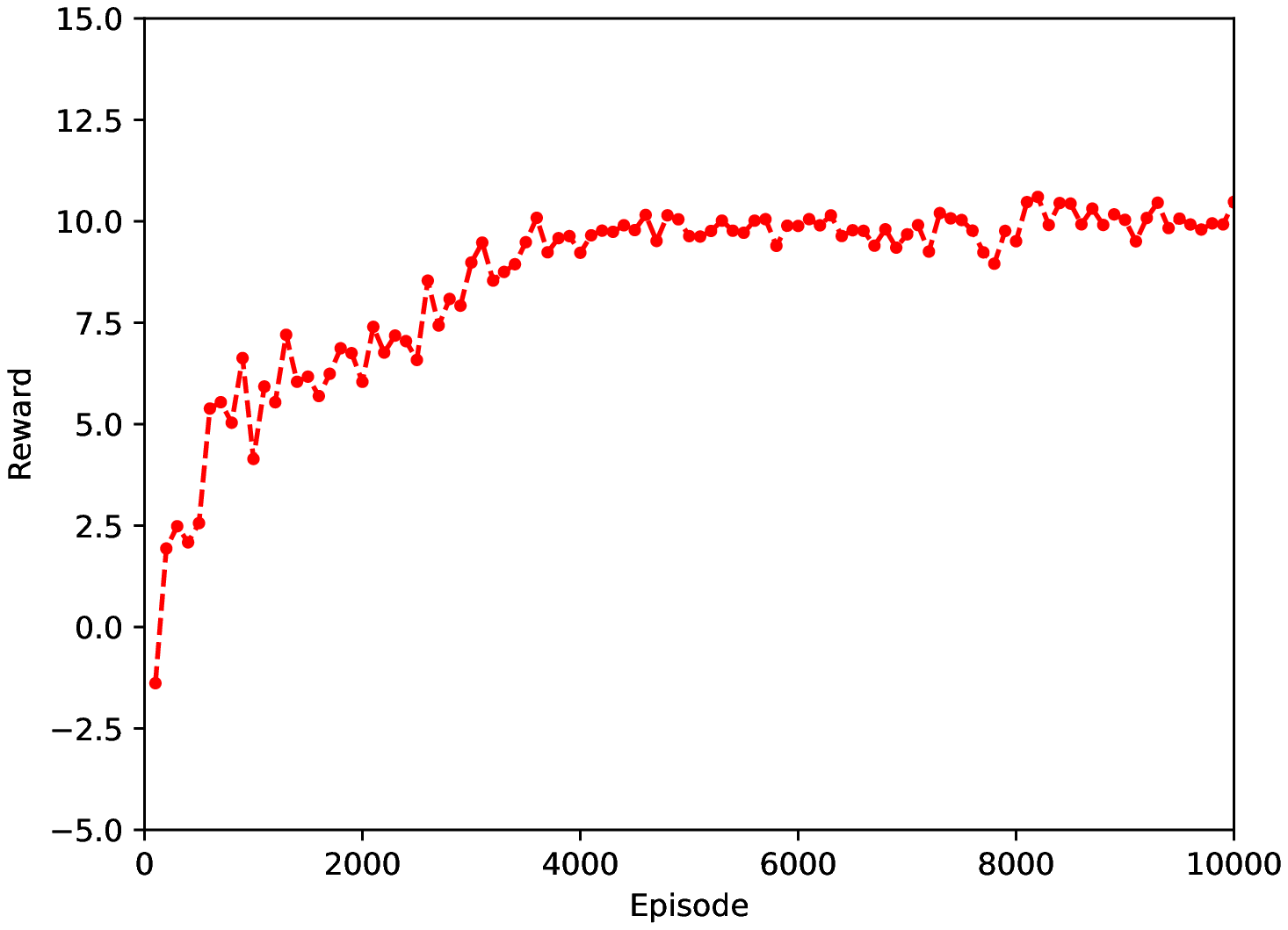}\\
		\vspace{0.02cm}
		\includegraphics[width=1.0\linewidth]{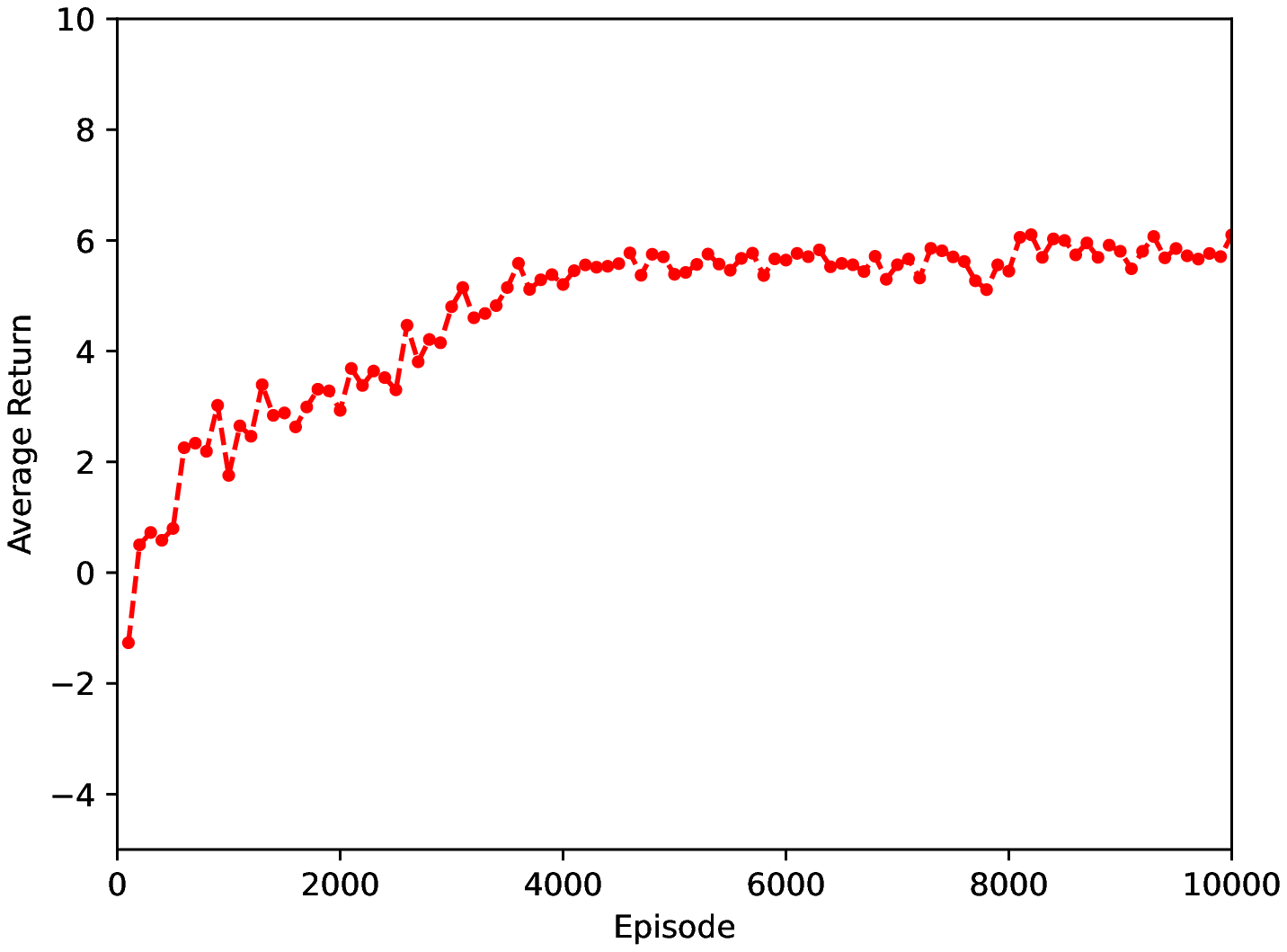}\\
		\vspace{0.02cm}
		\includegraphics[width=1.0\linewidth]{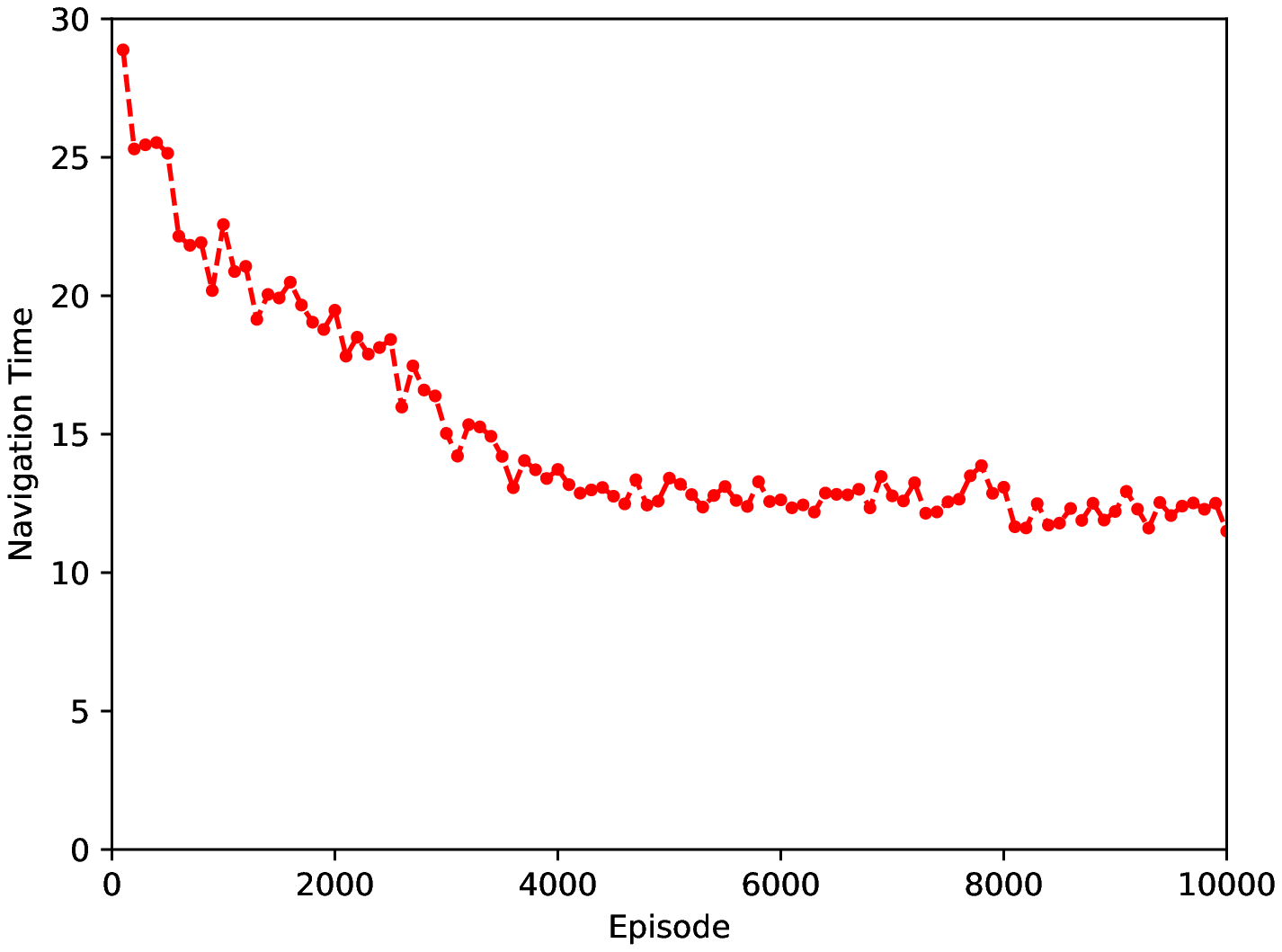}\\
        \vspace{0.02cm}
		\includegraphics[width=1.0\linewidth]{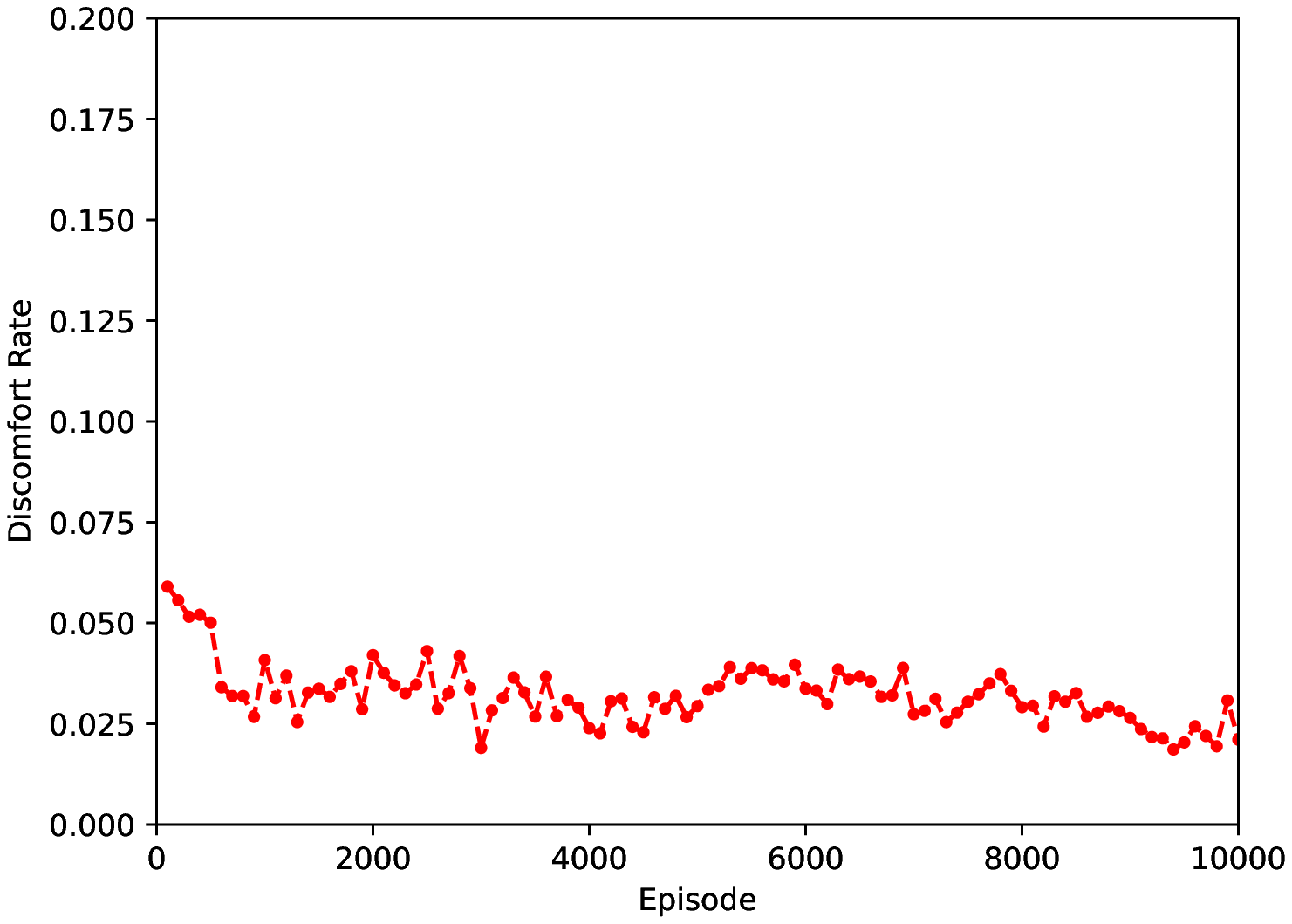}\\
		\label{5h_training}
        \vspace{0.1cm}
		\end{minipage}
	}
	\subfigure[Training in complex scenarios]{
	    \begin{minipage}[t]{0.46\linewidth}
	    \centering
		\includegraphics[width=1.0\linewidth]{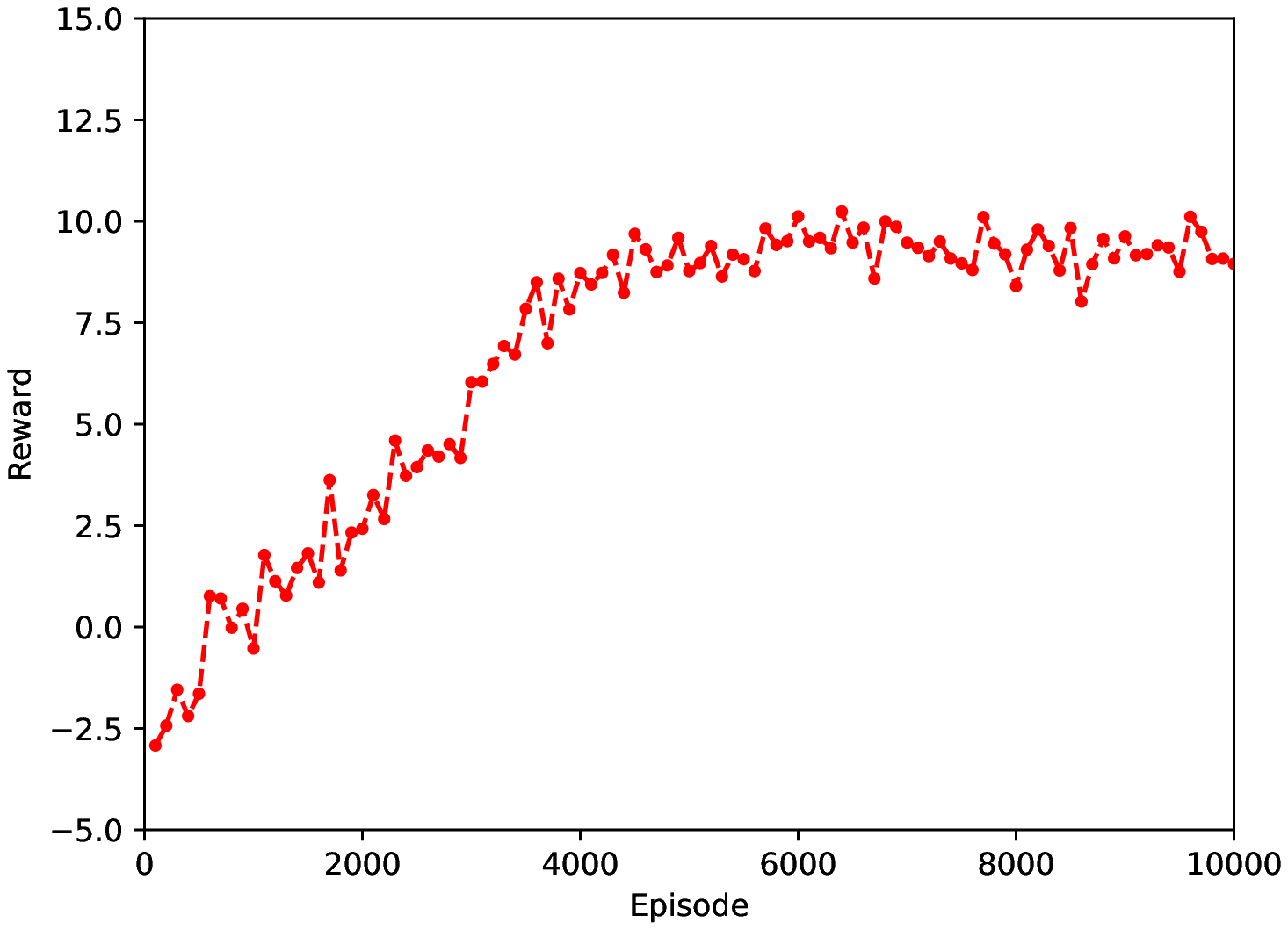}\\
		\vspace{0.02cm}
		\includegraphics[width=1.0\linewidth]{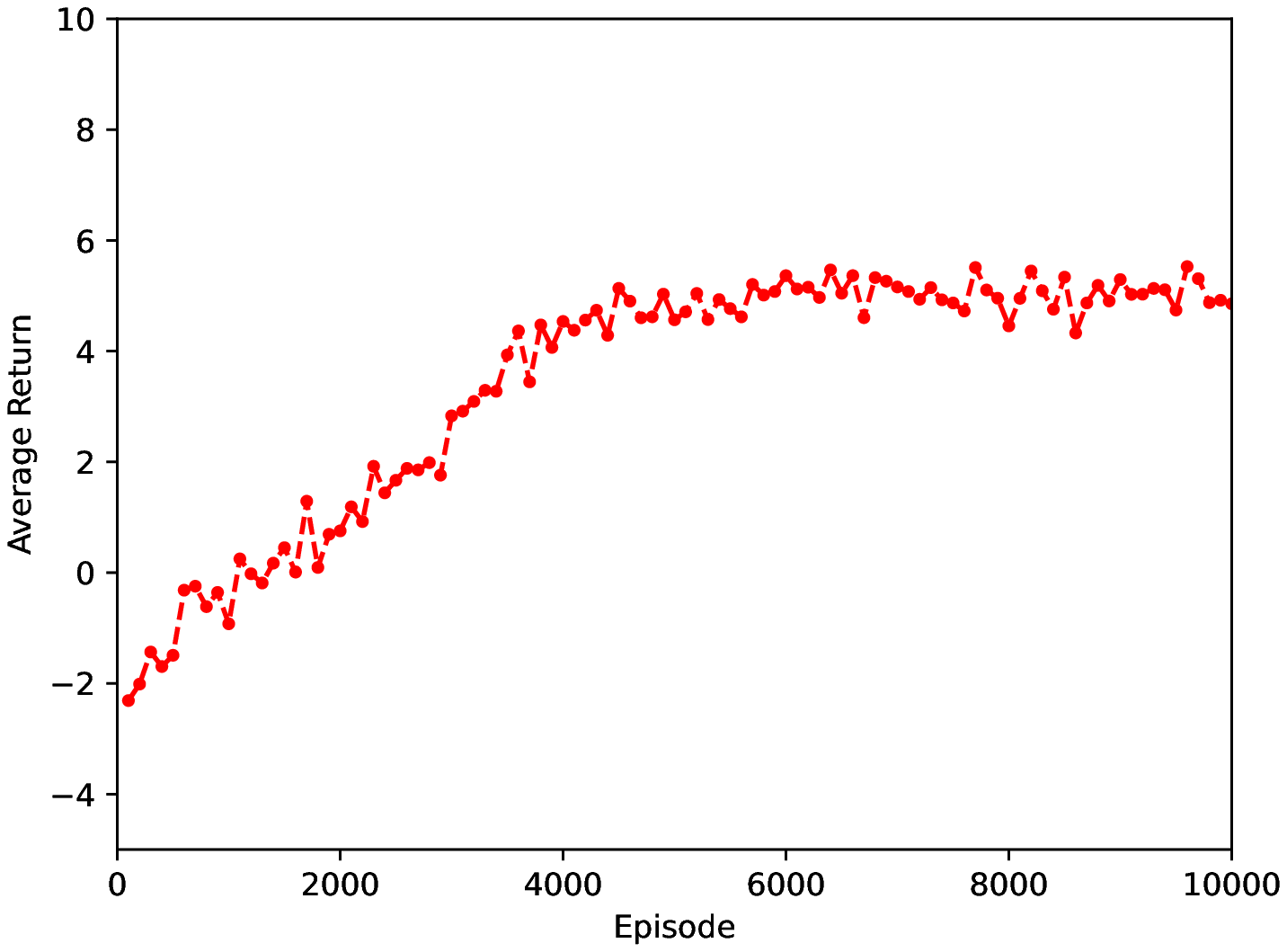}\\
		\vspace{0.02cm}
		\includegraphics[width=1.0\linewidth]{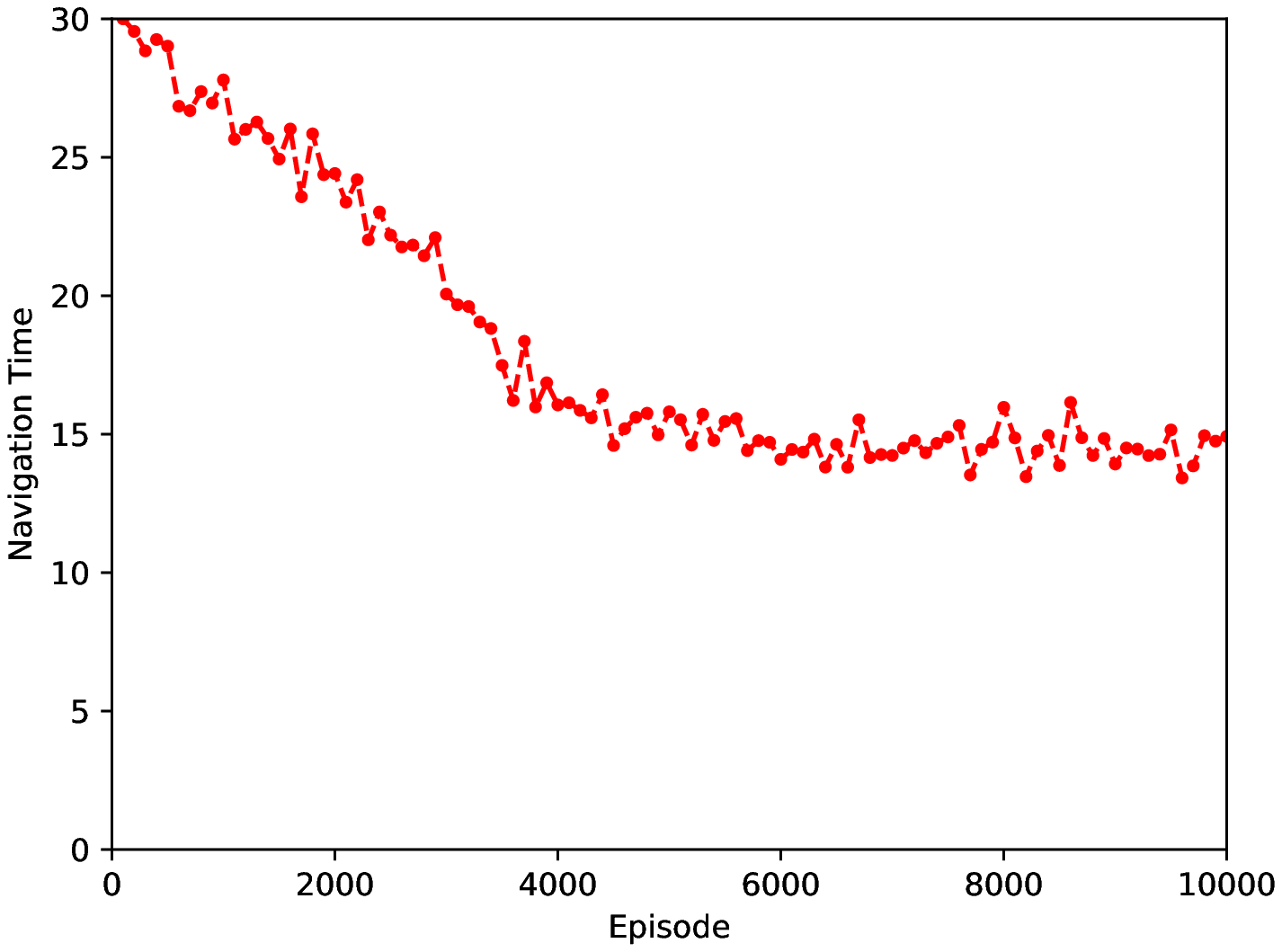}\\
        \vspace{0.02cm}
		\includegraphics[width=1.0\linewidth]{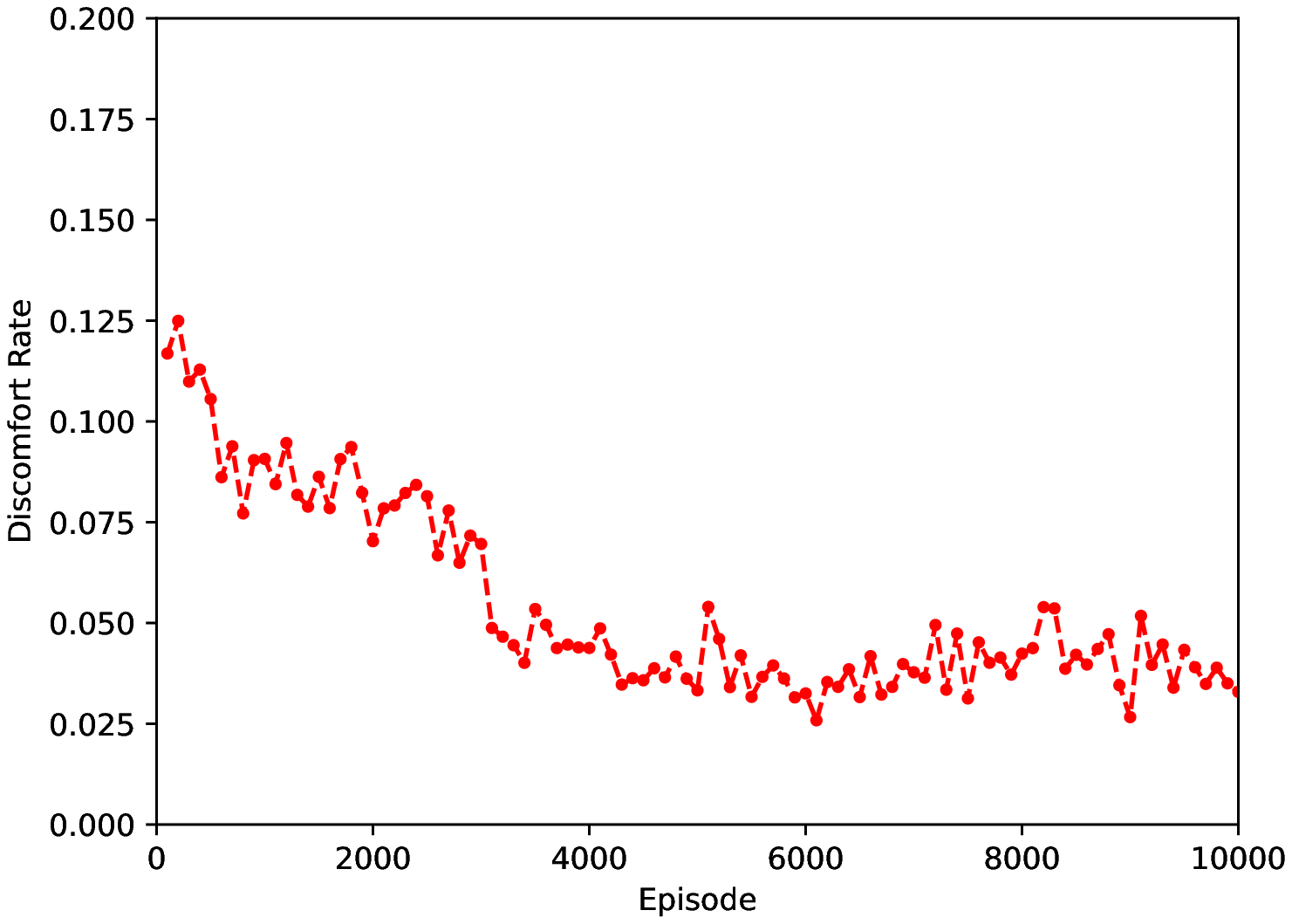}\\
		\label{10h_training}
		\vspace{0.1cm}
		\end{minipage}
	}
	\caption{Training curves in simple scenarios (Fig. \ref{5h_training}) and complex scenarios (Fig. \ref{10h_training}). From top to bottom, the y-axis corresponds to the total reward, average return, navigation time and discomfort rate per episode, averaged over the last 100 episodes.} 
	\begin{spacing}{1.0}
	\end{spacing}
	\label{training}
	\begin{spacing}{0.0}
\end{spacing}
\end{figure}

SG-DQN is trained in both simple scenarios and complex scenarios. The resulting training curves are shown in Fig. \ref{training}. At the beginning of training, it is difficult to complete the crowd navigation task with a randomly initialized model, and most termination states are timeout. As the training continues, the robot quickly learns to keep a safe distance from pedestrians and slowly understands the crowd. In the last period of training, SG-DQN achieves relatively stable performance. The difference between different scenarios is predictable. The main reason is that there are more interactions in complex scenarios, which makes the environment more challenging and difficult. More detailed quantitative results are described in Sec. \ref{quantitative}.
\subsubsection{Collision Avoidance Behaviors} 
\begin{figure}[htbp] 
	\subfigure[Trajectory in a simple scenario]{
	\includegraphics[width=0.45\linewidth ,height=1.5in]{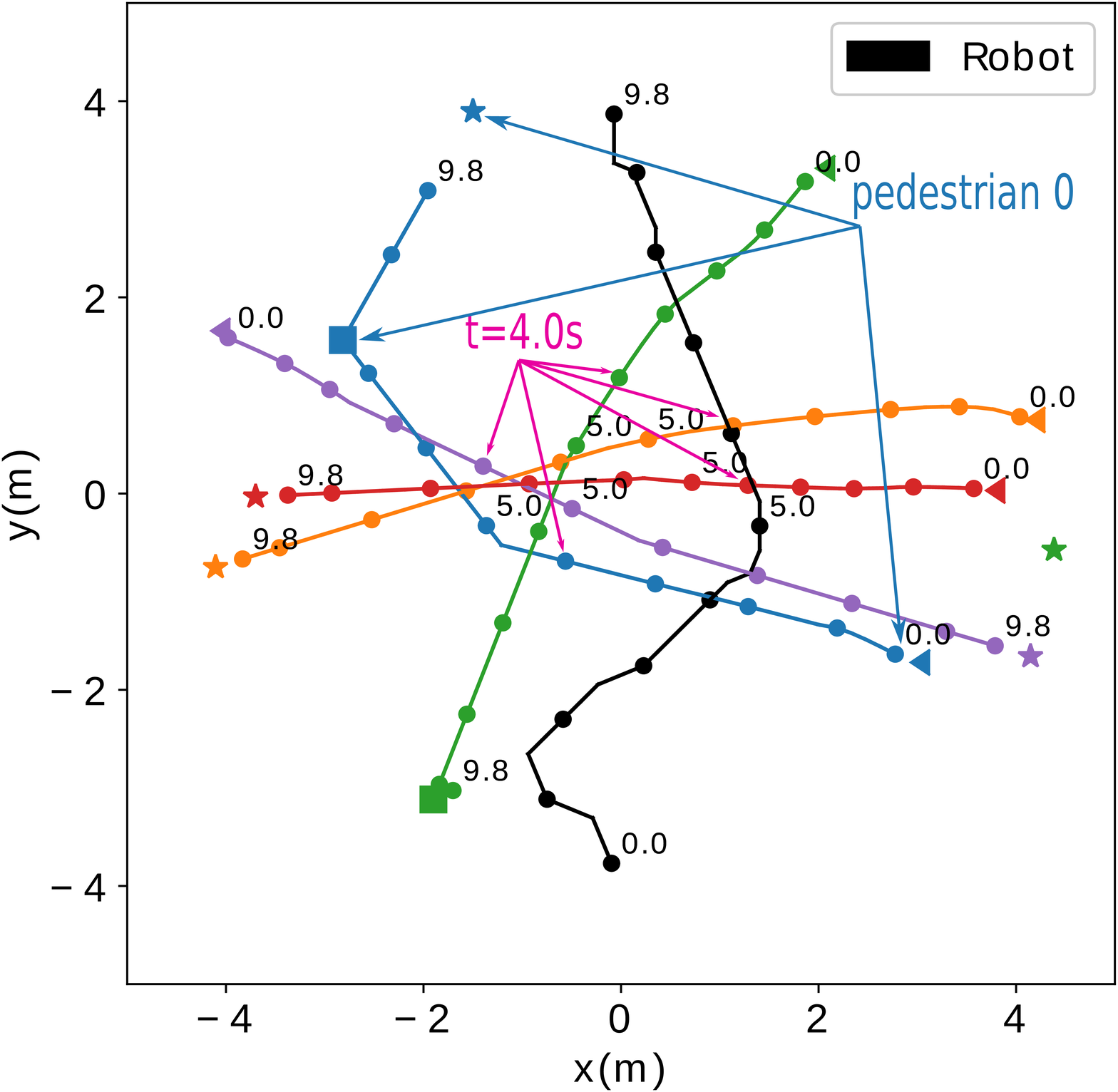}
	\label{sim_traj}}
	\subfigure[Trajectory in a complex scenario]{
	\includegraphics[width=0.45\linewidth, height=1.5in]{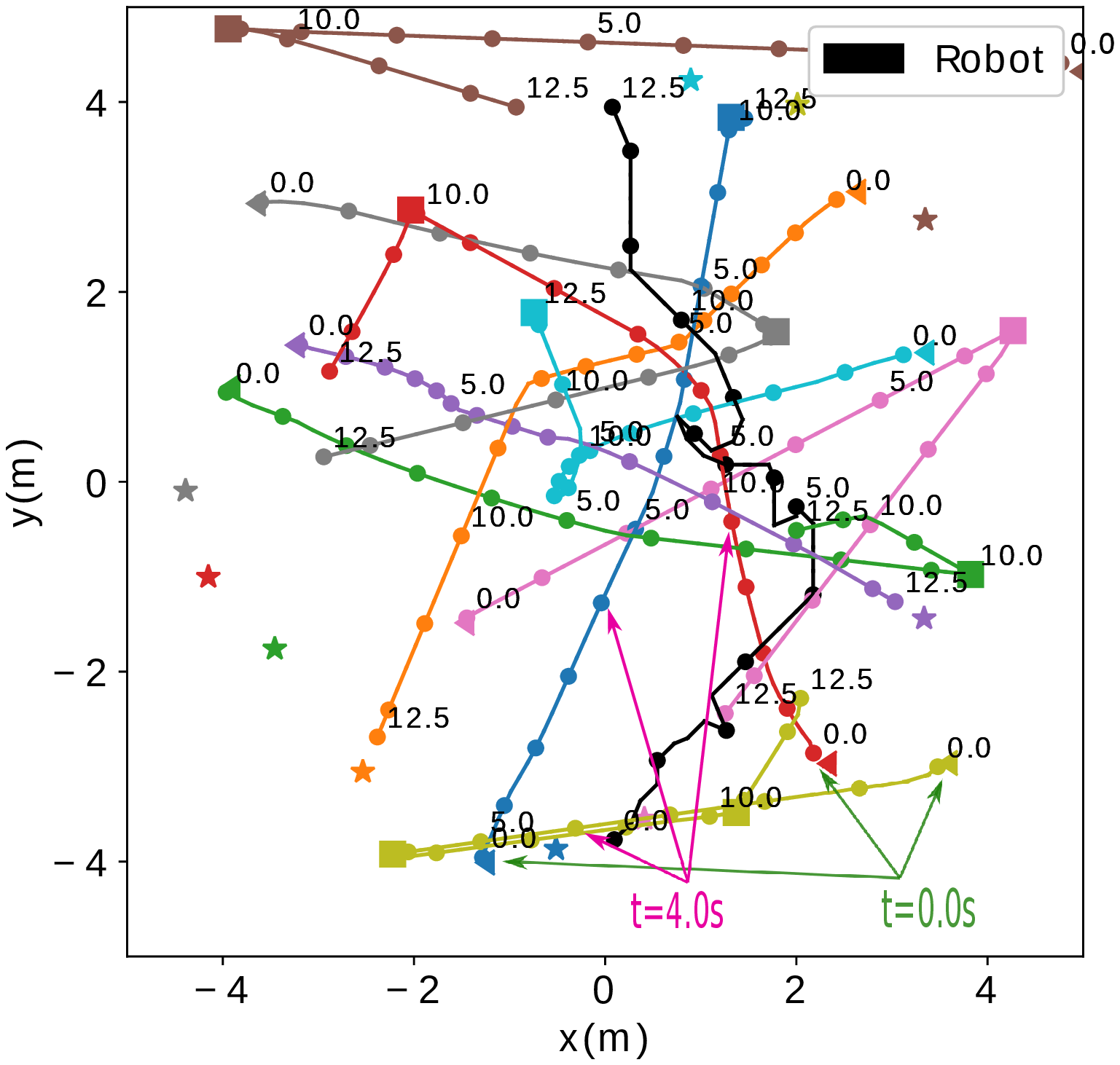}
	\label{com_traj}}
	\caption{Trajectory diagrams for a simple scenario (Fig. \ref{sim_traj}) and a complex scenario (Fig. \ref{com_traj}). Here, the discs represent agents, black for the robot and other colors for pedestrians. The numbers near the discs indicate the time. The time interval between two consecutive discs is 1.0 $s$. Here, the initial positions, the turning positions and the final goal positions are marked with triangles, squares and  five-pointed stars, respectively.} 
\label{traj}
\end{figure}
With the learned policy, the robot is able to reach its goal position safely and quickly in both simple and complex scenarios. The resulting trajectory diagrams are shown in Fig \ref{traj}. In the complex scenario, the robot has to pay more attention on avoiding pedestrians, resulting a more rought trajectory and a longer navigation time. In both simple and complex scenarios, the robot performs proactive and foresighted collision avoidance behaviors. The robot can always recognize and avoid the approaching interaction center of the crowd. For example, in the simple scenario, the robot turns right suddenly at approximately 1.5 $s$ to avoid the potential encirclement at 4.0 $s$. In addition, as shown in Fig. \ref{com_traj}, even if the robot is trapped in an encirclement of pedestrians, it has the ability to escape from the environment safely. Here, the encirclement of three pedestrians starts at 0.0 $s$ and lasts approximately 4.0 $s$. The encirclements are indicated by lines with arrows.
\subsubsection{Attention Modeling} 
\begin{figure}[tbp] 
	\subfigure[Simple scenario]{
	\includegraphics[width=0.46\linewidth]{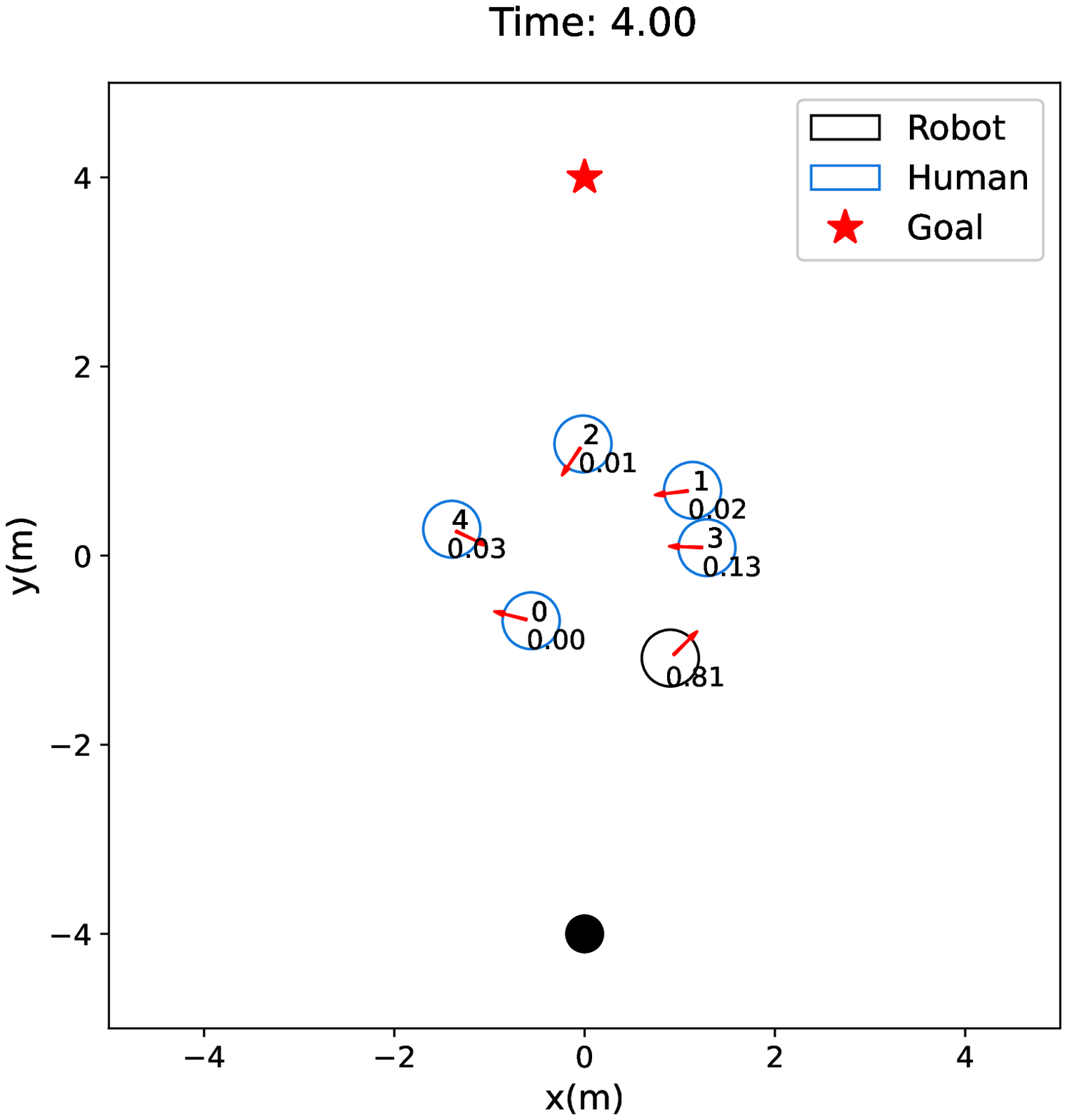}
	\label{sim_att_wei}}
	\subfigure[Complex scenario]{
	\includegraphics[width=0.46\linewidth]{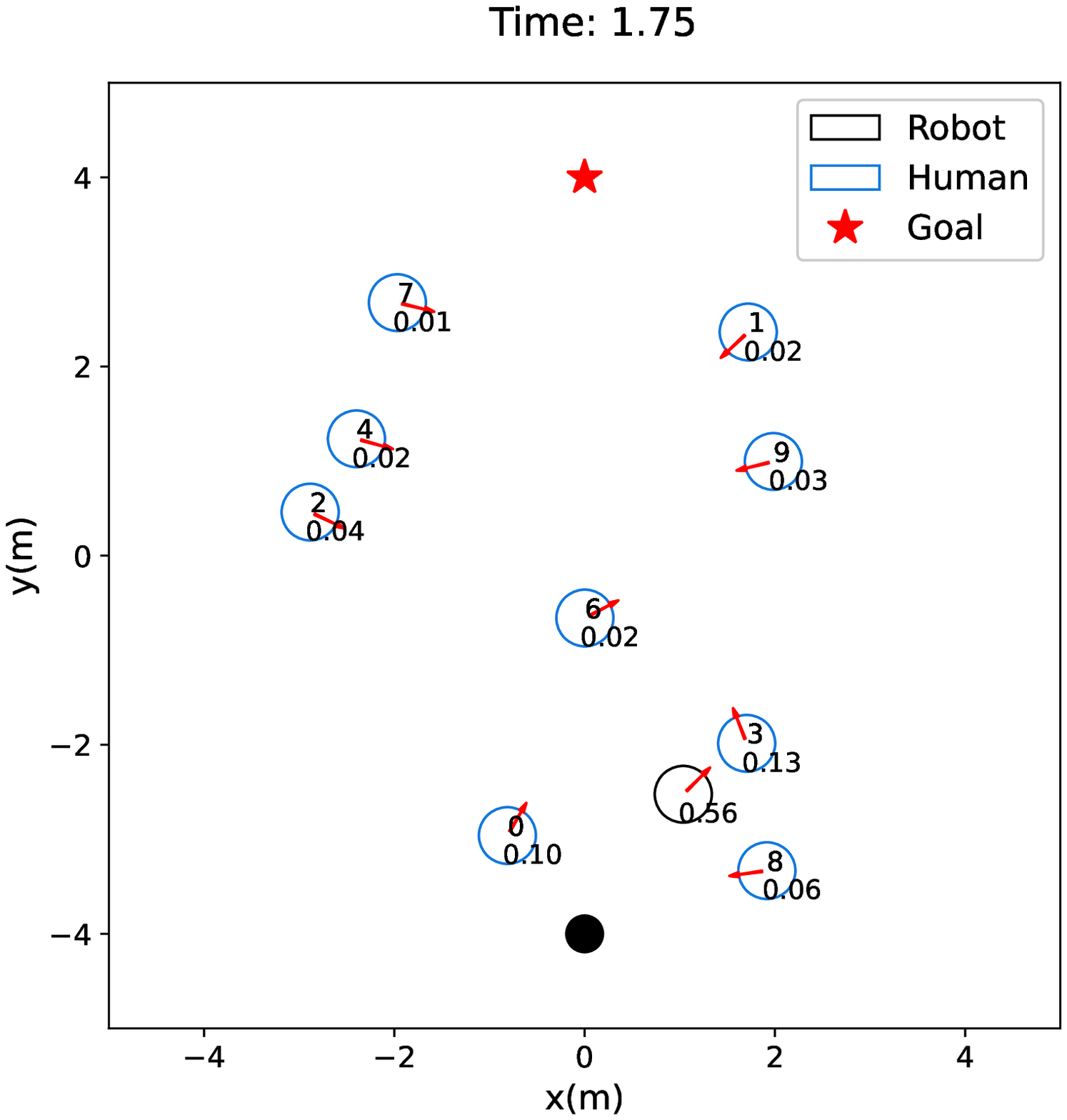}
	\label{com_att_wei}}
	\caption{Robot attention weights for agents in a simple scenario (Fig. \ref{sim_att_wei}) and a complex scenario (Fig. \ref{com_att_wei}). The natural numbers are the serial numbers and the two-digit decimals are the attention weights the robot gives to agents. The red arrows indicate the velocity attitudes of agents. The initial position and goal position of the robot are marked by a black disc and a red five-pointed star, respectively.}
	\label{att_wei}
	\begin{spacing}{-0.5}
	\end{spacing}
\end{figure}

Fig. \ref{att_wei} shows the attention weights in two test crowd scenes. In both simple and complex scenarios, the robot pays more attention to pedestrians who are close or moving towards it, e.g., pedestrian 3 in Fig. \ref{sim_att_wei} and pedestrians 0, 3, and 8 in Fig. \ref{com_att_wei}. Additionally, the robot also pays more attention to pedestrians who may interact with it. An example is the attention weight given to pedestrian 2 in the complex scenario, which shows that the robot's understanding of the crowd is foresighted. Another interesting point is the self-attention weight of the robot. The fewer pedestrians there are around the robot, the greater the self-attention weight. This means that the robot is able to balance its navigation task and collision avoidance behavior.\\
\subsection{Quantitative Evaluation}
\label{quantitative}
\indent Three existing state-of-the-art mothods, ORCA \cite{ORCA_van_2011}, LM-SARL\cite{cri_chen_2019} and MP-RGL-Onestep\cite{rgl_changan_2020}, are implemented as baseline methods. In addition, to verify the effect of the rollout performance, a dueling DQN version of SG-DQN is also developed as a contrast algorithm by setting the planning depth to 0. In the implementation of ORCA, the pedestrians' radii are set by $d_{s}=0.2$ to maintain a safe distance from pedestrians. For a fair comparison, RL methods apply the same  reward function. The implementation of MP-RGL-Onestep is different from the original version proposed in \cite{rgl_changan_2020}. With two independent graph models, the state value prediction and the human motion prediction are two separated modules. In addition, the training process has been repeated for six times and the resulting models are evaluated with 1000 random test cases. The random seeds in the 6 training processes are 7, 17, 27,37, 47, and 57. Finally, the statistical results are shown in Table \ref{simple_com} (for simple scenarios) and Table \ref{complex_com} (for complex scenarios).\\
\indent In the quantitative evaluation, the metrics includes: \textit{"Success"}, the success rate of robot reaching its goal safely; \textit{"Collision"}, the rate of the robot colliding with pedestrians; \textit{"Nav. Time"}, the navigation time to reach the goal in seconds; \textit{"Disc. Rate"}, the ratio of the number of steps cause discomfort (the robot violates the safety distance of the pedestrians) to the total number of steps; \textit{"Return"}, the discounted cumulative return averaged over stpes;  and \textit{"Run. Time"}, the running time per iteration in milliseconds. Here, all metrics are averaged over steps across all episodes and test cases.\\
\begin{table*}[htbp]
\begin{spacing}{1.0}
\end{spacing}
\caption{Quantitative results in simple secnarios. }
\begin{center}
\begin{spacing}{0.5}
\end{spacing}
\begin{tabular}{|c|c|c|c|c|c|c|}
\hline
\textbf{Methods} & \textbf{Success} & \textbf{Collision} & \textbf{Nav. Time(s)} &\textbf{Disc. Rate} &\textbf{Avg. Return} &\textbf{Run Time(ms)}\\
\hline
ORCA\cite{ORCA_van_2011} & 0.824 & 0.176 & 12.07 & 0.053 & 4.476  & \textbf{0.05} \\
\hline
dueling DQN & 0.986	& 0.013	& 11.25 & 0.011 &	6.120 & 2.67 \\
\hline
LM-SARL\cite{cri_chen_2019} & \textbf{0.997} & \textbf{0.003} & 10.95 & \textbf{0.013}  & 6.297  &42.50 \\
\hline
MP-RGL-Onestep\cite{rgl_changan_2020}& 0.997 & 0.003 & 10.57 & 0.017 &6.388  &23.07 \\
\hline
SG-DQN & 0.996 & 0.004 & \textbf{10.15} & 0.025 &\textbf{6.507}  & 10.18 \\
\hline
\end{tabular}
\begin{spacing}{0.0}
\end{spacing}
\begin{spacing}{0.5}
\end{spacing}
\label{simple_com}
\end{center}
\end{table*}
\begin{table*}[htbp]
\begin{spacing}{2.0}
\end{spacing}
\caption{Quantitative results in complex secnarios. }
\begin{center}
\begin{spacing}{0.5}
\end{spacing}
\begin{tabular}{|c|c|c|c|c|c|c|}
\hline
\textbf{Methods} & \textbf{Success} & \textbf{Collision} & \textbf{Nav. Time(s)} &\textbf{Disc. Rate} &\textbf{Avg. Return}  &\textbf{Run Time(ms)}\\
\hline
ORCA\cite{ORCA_van_2011} & 0.769 & 0.222 & 13.88 & 0.095 & 3.689 & \textbf{0.08} \\
\hline
dueling DQN & 0.950 & 0.048 & 13.66 & 0.029 & 5.283  & 2.76\\
\hline
LM-SARL\cite{cri_chen_2019} & 0.993 & 0.007 & 12.47 & \textbf{0.015} & 5.860  &53.10 \\
\hline
MP-RGL-Onestep\cite{rgl_changan_2020}& \textbf{0.994} & \textbf{0.006} & 11.88 & 0.028 & 6.033  &29.04 \\
\hline
SG-DQN & 0.992 & 0.008 & \textbf{11.71} & 0.032 & \textbf{6.035}  &14.12  \\
\hline
\end{tabular}
\begin{spacing}{0.0}
\end{spacing}
\begin{spacing}{0.5}
\end{spacing}
\label{complex_com}
\end{center}
\end{table*}
\indent As expected, the ORCA method has the highest collision rate, the longest navigation time and the best real-time performance in both simple and complex scenarios. This is because that the ORCA method is a totally reactive policy, which easily induces the robot to fall into the \textit{inevitable collision states} (ICS) \cite{ics_fraichard_2004} and causes the freezing robot problem\cite{freeze_trautman_2010}. The results of ORCA show the necessity of a foresighted policy.\\
\indent The dueling DQN acheives better performance than ORCA but worse performance than the other three RL algorithms. On the one hand, this shows that even though the dueling DQN is a reactive policy, it can help the robot to make foresighted decisions after being trained with much experience. On the other hand, it illustrates that as the scenario becomes increasingly complex it will be increasingly challenging to learn a simple control policy mapping the raw state to the best option directly. It is necessary to integrate the dueling DQN with online planning.\\
\indent All of LM-SARL, MP-RGL-Onestep and SG-DQN achieve performance far superior to that of dueling DQN, with success rates higger than 0.99. In simple scenarios, SG-DQN performs better than all others, achieving the shortest navigation time and the largest discounted cumulative return. It can be attributed to the dueling DQN, which stores some shortcuts and is independent of the learned environment model. In complex scenarios, the performance of SG-DQN is at least equivalent to, if not better than, others. The increased crowd interactions requires the robot to pay more attention to avoiding pedestrians, and narrows the gap between SG-DQN and the other two algorithms. Next, let us turn our attention to the metric of \textit{Run. Time}. Regardless of whether the scenario is simple or complex, SG-DQN requires a much lower computational cost, taking 10.18 ms and 14.12 ms per iteration in simple and complex scenarios, respectively. It is approximately half of the time of MP-RGL-Onestep and a quater that of LM-SARL. In addition, SG-DQN causes a slight decrease in \textit{Success} and an slight increase in \textit{Disc. Rate}. Considering the halved computational cost, this is quite valuable.
\section{Conclusion}
\indent In this paper, we propose SG-DQN, a graph-based reinforcement learning method, for mobile robots in a crowd, with a high success rate of more than 0.99. Compared against the state-of-the-art methods, SG-DQN achieves equivalent, if not better, performance in both simple scenarios and complex scenarios, while requiring less than half of the computational cost. Its success can be attributed to three innovations proposed in this work. The first innovation is the introduction of a social attention mechanism in the spatial graph convolution operation. With the improved two-layer GAT, it is available to extract an efficient graph representation for the crowd-robot state. The second innovation is the application of a dueling DQN, which can directly evaluate coarse q-values of the current state and quickly generate the best candidate actions. It greatly reduces the computational cost. The third innovation is the integration of the learning method and online planning. By performing rollouts on the current state, the coarse q-values generated by the dueling DQN are refined with a tree search. These innovations may also be useful in other similar applications.
\bibliographystyle{hieeetr}
\bibliography{ref}

\begin{thebibliography}{10}

\bibitem{sfm_helbing_1995}
D.~Helbing and P.~Moln\'ar, ``Social force model for pedestrian dynamics,''
  {\em Phys. Rev. E}, vol.~51, pp.~4282--4286, May 1995.

\bibitem{sfm_ferrer_2013}
G.~Ferrer, A.~Garrell, and A.~Sanfeliu, ``Robot companion: A social-force based
  approach with human awareness-navigation in crowded environments,'' in {\em
  2013 IEEE/RSJ International Conference on Intelligent Robots and Systems},
  pp.~1688--1694, 2013.

\bibitem{vo_fiorini_1998}
P.~Fiorini and Z.~Shiller, ``Motion planning in dynamic environments using
  velocity obstacles,'' {\em The International Journal of Robotics Research},
  vol.~17, no.~7, pp.~760--772, 1998.

\bibitem{RVO_van_2008}
J.~van~den Berg, M.~Lin, and D.~Manocha, ``Reciprocal velocity obstacles for
  real-time multi-agent navigation,'' in {\em 2008 IEEE International
  Conference on Robotics and Automation (ICRA)}, pp.~1928--1935, 2008.

\bibitem{ORCA_van_2011}
J.~van~den Berg, G.~S.J., M.~Lin, and D.~Manocha, ``Reciprocal n-body collision
  avoidance,'' in {\em Robotics Research}, pp.~3--19, 2011.

\bibitem{brvo_kim_2015}
S.~Kim, S.~J. Guy, W.~Liu, D.~Wilkie, R.~W. Lau, M.~C. Lin, and D.~Manocha,
  ``Brvo: Predicting pedestrian trajectories using velocity-space reasoning,''
  {\em The International Journal of Robotics Research}, vol.~34, no.~2,
  pp.~201--217, 2015.

\bibitem{social_2016_alahi}
A.~Alahi, K.~Goel, V.~Ramanathan, A.~Robicquet, L.~Fei-Fei, and S.~Savarese,
  ``Social lstm: Human trajectory prediction in crowded spaces,'' in {\em 2016
  IEEE Conference on Computer Vision and Pattern Recognition (CVPR)},
  pp.~961--971, 2016.

\bibitem{social_2018_vemula}
A.~Vemula, K.~Muelling, and J.~Oh, ``Social attention: Modeling attention in
  human crowds,'' in {\em 2018 IEEE International Conference on Robotics and
  Automation (ICRA)}, pp.~4601--4607, 2018.

\bibitem{survey_Rudenko_2019}
A.~Rudenko, L.~Palmieri, S.~Herman, K.~M. Kitani, D.~M. Gavrila, and K.~Arras,
  ``Human motion trajectory prediction: a survey,'' {\em The International
  Journal of Robotics Research}, vol.~39, pp.~895--935, 2019.

\bibitem{predict_katyal_2020}
K.~D. Katyal, G.~D. Hager, and C.~M. Huang, ``Intent-aware pedestrian
  prediction for adaptive crowd navigation,'' in {\em 2020 IEEE International
  Conference on Robotics and Automation (ICRA)}, pp.~3277--3283, 2020.

\bibitem{social_sun_2020}
J.~Sun, Q.~Jiang, and C.~Lu, ``Recursive social behavior graph for trajectory
  prediction,'' in {\em 2020 IEEE/CVF Conference on Computer Vision and Pattern
  Recognition (CVPR)}, pp.~657--666, 2020.

\bibitem{trajectory_tim_2020}
T.~Salzmann, B.~Ivanovic, P.~Chakravarty, and M.~Pavone, ``Trajectron++:
  Dynamically-feasible trajectory forecasting with heterogeneous data,'' 2021,
  arXiv:2001.03093.

\bibitem{freeze_trautman_2010}
P.~Trautman and A.~Krause, ``Unfreezing the robot: Navigation in dense,
  interacting crowds,'' in {\em 2010 IEEE/RSJ International Conference on
  Intelligent Robots and Systems}, pp.~797--803, 2010.

\bibitem{chen_decen_2016}
Y.~F. Chen, M.~Liu, M.~Everett, and J.~P. How, ``Decentralized
  non-communicating multiagent collision avoidance with deep reinforcement
  learning,'' in {\em 2017 IEEE International Conference on Robotics and
  Automation (ICRA)}, pp.~285--292, 2017.

\bibitem{chen_socially_2017}
Y.~F. Chen, M.~Liu, M.~Everett, and J.~P. How, ``Socially aware motion planning
  with deep reinforcement learning,'' in {\em 2017 IEEE/RSJ International
  Conference on Intelligent Robots and Systems (IROS)}, pp.~1343--1350, 2017.

\bibitem{cri_chen_2019}
C.~Chen, Y.~Liu, S.~Kreiss, and A.~Alahi, ``Crowd-robot interaction:
  Crowd-aware robot navigation with attention-based deep reinforcement
  learning,'' in {\em 2019 International Conference on Robotics and Automation
  (ICRA)}, pp.~6015--6022, 2019.

\bibitem{yuying_navigation_2020}
Y.~Chen, C.~Liu, B.~E. Shi, and M.~Liu, ``Robot navigation in crowds by graph
  convolutional networks with attention learned from human gaze,'' {\em IEEE
  Robotics and Automation Letters}, vol.~5, no.~2, pp.~2754--2761, 2020.

\bibitem{rgl_changan_2020}
C.~Chen, S.~Hu, P.~Nikdel, G.~Mori, and M.~Savva, ``Relational graph learning
  for crowd navigation,'' in {\em 2020 IEEE/RSJ International Conference on
  Intelligent Robots and Systems (IROS)}, pp.~10007--10013, 2020.

\bibitem{mnih-atari-2013}
V.~Mnih, K.~Kavukcuoglu, D.~Silver, A.~Graves, I.~Antonoglou, D.~Wierstra, and
  M.~Riedmiller, ``Playing atari with deep reinforcement learning,'' in {\em
  NIPS Deep Learning Workshop}, 2013.

\bibitem{mnih-dqn-2015}
V.~Mnih, K.~Kavukcuoglu, D.~Silver, A.~A. Rusu, J.~Veness, M.~G. Bellemare,
  A.~Graves, M.~Riedmiller, A.~K. Fidjeland, G.~Ostrovski, S.~Petersen,
  C.~Beattie, A.~Sadik, I.~Antonoglou, H.~King, D.~Kumaran, D.~Wierstra,
  S.~Legg, and D.~Hassabis, ``Human-level control through deep reinforcement
  learning,'' {\em Nature}, vol.~518, no.~7540, pp.~529--533, 2015.

\bibitem{densecavoid_sathy_2020}
A.~J. Sathyamoorthy, J.~Liang, U.~Patel, T.~Guan, R.~Chandra, and D.~Manocha,
  ``Densecavoid: Real-time navigation in dense crowds using anticipatory
  behaviors,'' in {\em 2020 IEEE International Conference on Robotics and
  Automation (ICRA)}, pp.~11345--11352, 2020.

\bibitem{long_optimally_2018}
P.~Long, T.~Fan, X.~Liao, W.~Liu, H.~Zhang, and J.~Pan, ``Towards optimally
  decentralized multi-robot collision avoidance via deep reinforcement
  learning,'' in {\em 2018 IEEE International Conference on Robotics and
  Automation (ICRA)}, pp.~6252--6259, 2018.

\bibitem{Everett_DRL_2018}
M.~Everett, Y.~F. Chen, and J.~P. How, ``Motion planning among dynamic,
  decision-making agents with deep reinforcement learning,'' in {\em 2018
  IEEE/RSJ International Conference on Intelligent Robots and Systems (IROS)},
  pp.~3052--3059, 2018.

\bibitem{everett_collision_2020}
M.~Everett, Y.~F. Chen, and J.~P. How, ``Collision avoidance in pedestrian-rich
  environments with deep reinforcement learning,'' {\em IEEE Access}, vol.~9,
  pp.~10357--10377, 2021.

\bibitem{socialgan_gupta_2018}
A.~Gupta, J.~Johnson, L.~Fei-Fei, S.~Savarese, and A.~Alahi, ``Social gan:
  Socially acceptable trajectories with generative adversarial networks,'' in
  {\em 2018 IEEE/CVF Conference on Computer Vision and Pattern Recognition},
  pp.~2255--2264, 2018.

\bibitem{gat_velickovic_2018}
P.~Velickovic, G.~Cucurull, A.~Casanova, A.~Romero, P.~Li\`{o}, and Y.~Bengio,
  ``Graph attention networks,'' in {\em 6th International Conference on
  Learning Representations}, 2018.

\bibitem{dueling_wang_2016}
Z.~Wang, T.~Schaul, M.~Hessel, H.~Hasselt, M.~Lanctot, and N.~Freitas,
  ``Dueling network architectures for deep reinforcement learning,'' in {\em
  Proceedings of The 33rd International Conference on Machine Learning},
  vol.~48, pp.~1995--2003, 2016.

\bibitem{Lin_93}
L.~J. Lin, {\em Reinforcement Learning for Robots Using Neural Networks}.
\newblock PhD thesis, Carnegie Mellon University, Pittsburgh, January 1993.

\bibitem{oh_nips_2015}
J.~Oh, X.~Guo, H.~Lee, R.~Lewis, and S.~Singh, ``Action-conditional video
  prediction using deep networks in atari games,'' in {\em Proceedings of the
  28th International Conference on Neural Information Processing Systems},
  vol.~2, pp.~2863--2871, 2015.

\bibitem{oh_value_2017}
J.~Oh, S.~Singh, and H.~Lee, ``Value prediction network,'' in {\em Advances in
  Neural Information Processing Systems 30: Annual Conference on Neural
  Information Processing Systems}, pp.~6118--6128, 2017.

\bibitem{silver2017mastering}
D.~Silver, J.~Schrittwieser, K.~Simonyan, I.~Antonoglou, A.~Huang, A.~Guez,
  T.~Hubert, L.~Baker, M.~Lai, A.~Bolton, Y.~Chen, T.~Lillicrap, F.~Hui,
  L.~Sifre, G.~van~den Driessche, T.~Graepel, and D.~Hassabis, ``Mastering the
  game of go without human knowledge,'' {\em Nature}, vol.~550, pp.~354--359,
  2017.

\bibitem{schrittwieser_mastering_2020}
J.~Schrittwieser, I.~Antonoglou, T.~Hubert, K.~Simonyan, L.~Sifre, S.~Schmitt,
  A.~Guez, E.~Lockhart, D.~Hassabis, T.~Graepel, T.~Lillicrap, and D.~Silver,
  ``Mastering atari, go, chess and shogi by planning with a learned model,''
  {\em Nature}, vol.~588, pp.~604--609, 2020.

\bibitem{adam_2015}
D.~P. Kingma and J.~Ba, ``Adam: A method for stochastic optimization,'' in {\em
  3rd International Conference on Learning Representations}, 2015.

\bibitem{ics_fraichard_2004}
T.~Fraichard and H.~Asama, ``Inevitable collision states — a step towards
  safer robots?,'' {\em Advanced Robotics}, vol.~18, pp.~1001--1024, 2004.

\end{thebibliography}
\end{document}